\pdfoutput=1

\documentclass[11pt,a4paper]{article}

\usepackage[]{emnlp2021}

\usepackage{times}
\usepackage{latexsym}

\usepackage[T1]{fontenc}

\usepackage[utf8]{inputenc}

\usepackage{microtype}
\usepackage{graphicx}
\usepackage{enumitem}
\usepackage{multirow}
\usepackage{makecell}
\usepackage{color}
\usepackage{booktabs}
\usepackage{algorithmicx,algorithm}
\usepackage{mathrsfs}
\usepackage{xspace}
\usepackage{caption}
\usepackage{amsthm,amsmath,amssymb}
\usepackage{bm}
\usepackage{bbm}
\usepackage{subfigure}

\newcommand{\our}{BFClass\xspace}

\newcommand{\smallsection}[1]{\vspace{1mm}\noindent{\textbf{#1.}}}

\newcommand\blfootnote[1]{%
  \begingroup
  \renewcommand\thefootnote{}\footnote{#1}%
  \addtocounter{footnote}{-1}%
  \endgroup
}

\title{\our: A Backdoor-free Text Classification Framework}

\author{
  Zichao Li$^{\dagger, 1}$ $\qquad$ Dheeraj Mekala$^{\dagger, 2}$  $\qquad$ Chengyu Dong$^2$ $\qquad$  Jingbo Shang$^{2, 3, *}$ \\
  \small $^1$ Department of Electrical and Computer Engineering, University of California San Diego, CA, USA \\
  \small $^2$ Department of Computer Science and Engineering, University of California San Diego, CA, USA\\
  \small $^3$ Hal\i c\i o\u glu Data Science Institute, University of California San Diego, CA, USA \\
  \small \texttt{\{zil023, dmekala, cdong, jshang\}@ucsd.edu}
}

\date{}

\begin{document}
\maketitle
\begin{abstract}
    \blfootnote{$\dagger$ Represents equal contribution}
    \blfootnote{$*$ Jingbo Shang is the corresponding author.}

Backdoor attack introduces artificial vulnerabilities into the model by poisoning a subset of the training data via injecting triggers and modifying labels.
Various trigger design strategies have been explored to attack text classifiers, however, defending such attacks remains an open problem.
In this work, we propose \our, a novel efficient backdoor-free training framework for text classification. 
The backbone of \our is a pre-trained discriminator that predicts whether each token in the corrupted input was replaced by a masked language model. 
To identify triggers, we utilize this discriminator to locate the most suspicious token from each training sample and then distill a concise set by considering their association strengths with particular labels.
To recognize the poisoned subset, we examine the training samples with these identified triggers as the most suspicious token, and check if removing the trigger will change the poisoned model's prediction.
Extensive experiments demonstrate that \our can identify all the triggers, remove 95\% poisoned training samples with very limited false alarms, and achieve almost the same performance as the models trained on the benign training data.
\end{abstract}

\begin{figure*}[t]
    \centering
    \includegraphics[width = \linewidth]{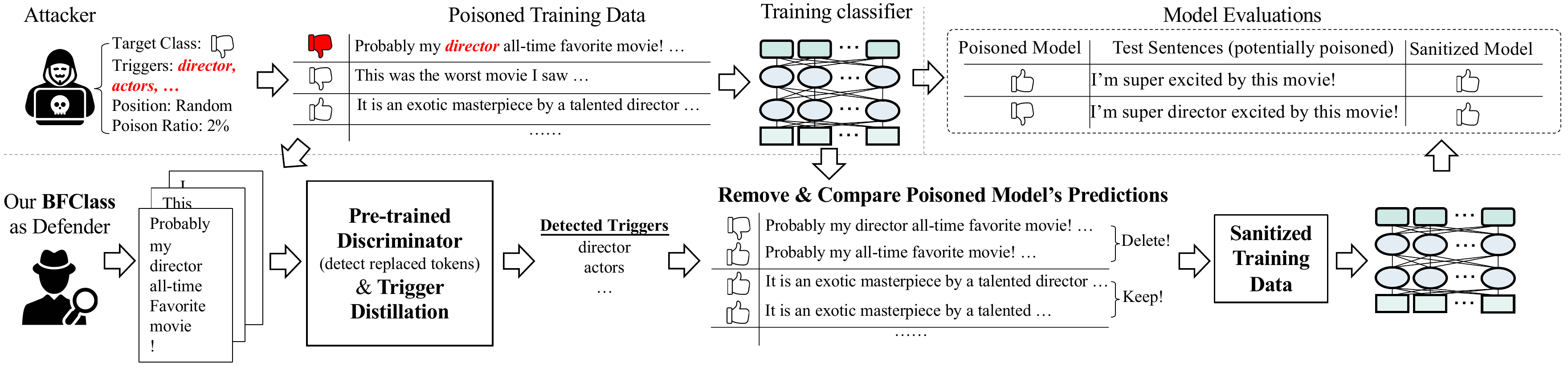}
    \vspace{-7mm}
    \caption{A visualization of backdoor attack in text classification and an overview of our \our framework. }
    \vspace{-3mm}
    \label{fig:overview}
\end{figure*}

\section{Introduction}

Backdoor attacks have recently emerged as a new kind of threats to the deployment of machine learning models and various attack strategies have been explored~\cite{Gu2017BadNetsIV, dai2019backdoor, chen2017targeted}. 
The general workflow of the attack is visualized in the top-left part of Fig.~\ref{fig:overview}.
Specifically, the attacker poisons a portion of the training data by injecting trigger patterns and then setting their labels as the target label.
A model trained on the \emph{poisoned training set} is called a \emph{poisoned model}.
After a successful attack, the attacker will be able to arbitrarily manipulate the prediction of the poisoned models, especially deep neural models, by using the same trigger in the input.
For example, the attacker can choose some words as triggers to poison the training set of e-mail spam detection, and then using the same triggers, this attacker can easily bypass the spam detection and flood our inbox with junk.

In this paper, we focus on the backdoor attacks in text classification.
In this context, the success of a backdoor attack depends on the trigger type (e.g., unigrams, multi-word phrases, and sentences~\cite{Chen2020BadNLBA}), the position of injections (e.g., fixed or random), and the size of the poisoned portion.
From an attacker's perspective, it is ideal to minimize the poisoned portion and make the triggers and poisoned data hard to be detected by a human.
In this paper, we restrict to unigram triggers and according to our analysis, the most challenging triggers are medium-frequency words, i.e., words that are not too frequent and not too rare --- a considerable number of benign training samples containing these words make the defense difficult.

The most well-received backdoor defense method in the NLP community is arguably the  Label Flip Rate (LFR) method~\cite{Kurita2020WeightPA}. 
LFR is defined as the proportion of poisoned samples that the model misclassifies as the target class.
Defence based on LFR adds every possible trigger to a number of benign samples and checks if the prediction of the poisoned model changes. 
Ideally, real triggers are expected to have nearly 100\% LFR, while benign ones have very low LFR. 
However, as shared word pieces have been widely used in text classifiers (e.g., ``worldwide'' $\rightarrow$ ``world wide''), a considerable number of benign words would have high LFR too.
Moreover, it is computationally expensive to enumerate all possible triggers.

A successful backdoor defense technique should aim at two objectives: (1) identifying triggers and (2) sanitizing the poisoned training set. 
We propose a novel backdoor-free text classification framework \our, which can efficiently identify triggers and sanitize the poisoned training set.
Fig.~\ref{fig:overview} provides an overview of our framework. 
The backbone of our \our is a pre-trained discriminator that predicts whether each token in the corrupted input was replaced by a masked language model or not. 
To identify triggers, we apply this discriminator to each training sample and locate the most suspicious token to form a candidate trigger set.
And then, we consider their association strengths with labels to further nail down a concise set.
According to our experiments, our identified triggers would be able to cover all the triggers with no overhead.
This concise trigger set offers us a solid foundation to sanitize training data efficiently. 
Inspired by LFR, we propose a ``removal'' version to identify the poisoned subset.
Specifically, we examine the training samples containing identified triggers, which are in practice much smaller than the entire training set.
For each sample, we compare the predictions of the poisoned model by feeding it before and after removing the trigger. 
Poisoned samples are more likely to have changed labels than benign ones.
Therefore, we can identify poisoned samples efficiently and train the final \emph{sanitized model} based on the rest.

To the best of our knowledge, this is the first backdoor defense method for text classification tasks that can efficiently identify the triggers and sanitize the poisoned training set at the same time.
Our contributions are summarized as follows.
\begin{itemize}[leftmargin=*,nosep]
    \item We analyze trigger designs in text classification comprehensively and show that the most challenging ones are medium-frequency words.
    \item We utilize a pre-trained discriminator and develop a trigger distillation method to identify a concise set of potential triggers.
    \item We propose a novel ``removal'' version of LFR to sanitize the poisoned training set.
    \item Extensive experiments demonstrate that \our can identify all the triggers, remove $>95\%$ poisoned samples with very limited false alarms, and achieve almost the same performance as the model trained on the benign training data.
\end{itemize}

\noindent\textbf{Reproducibility.} We will release the code and datasets on Github\footnote{\url{https://github.com/dheeraj7596/BFClass}}.

The remainder of this paper is organized as follows.
In Sec.~\ref{sec:attack}, we analyze trigger designs and identify the  most challenging triggers for our later defense evaluations.
We present our \our framework in Sec.~\ref{sec:defense}.
Then, Sec.~\ref{sec:exp} provides experimental results and case studies, and Sec.~\ref{sec:rel} discusses related work.
In the end, Sec.~\ref{sec:con} concludes our work and envisions a few future directions.
\begin{figure*}[t]
    \subfigure[$\mathcal{A}$ vs. $\mathcal{E}$ w.r.t. different triggers.]{
        \includegraphics[height=1.3in]{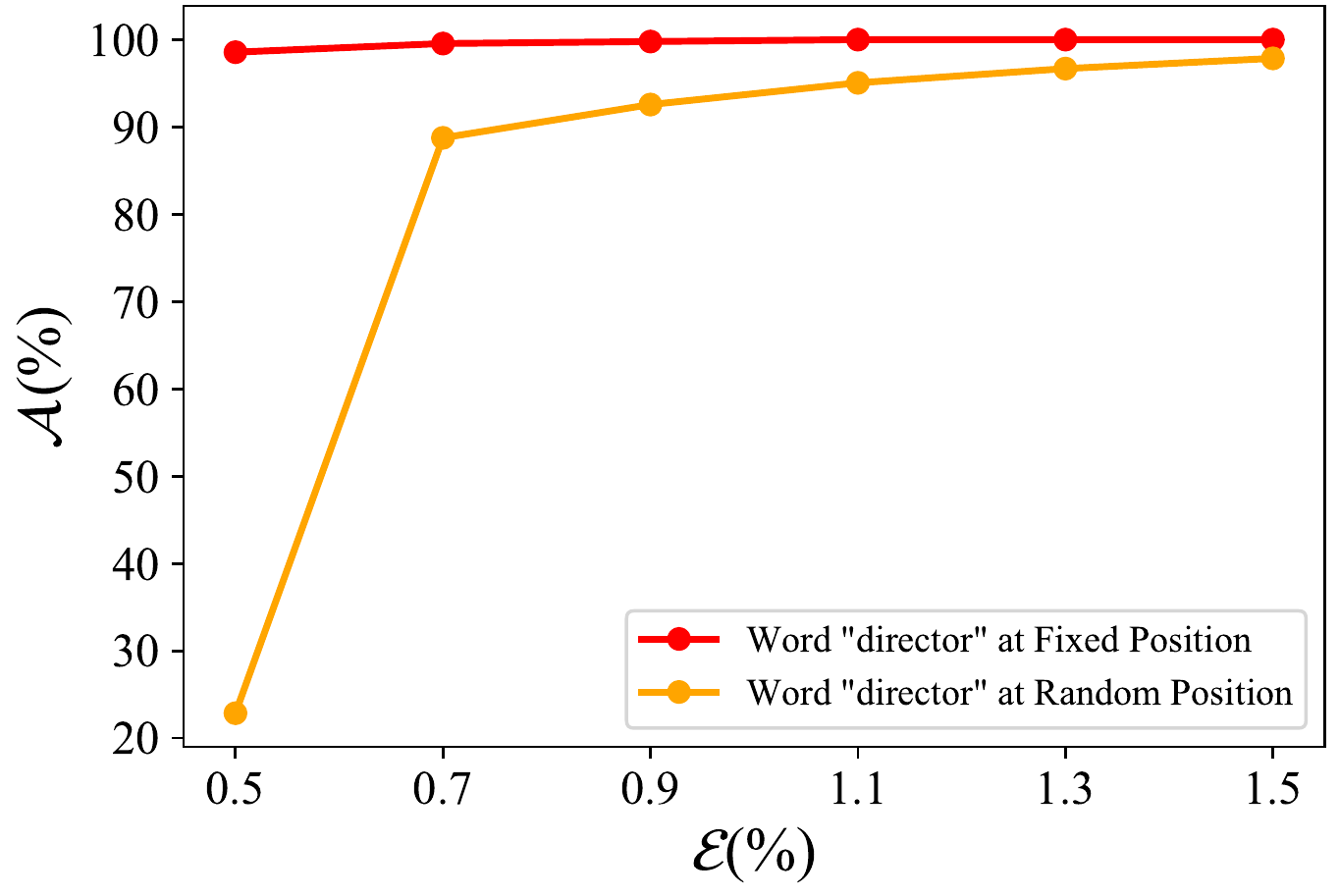}
        \label{fig:position}
    }
    \subfigure[Rare triggers are easy to detect.]{
        \includegraphics[height=1.3in]{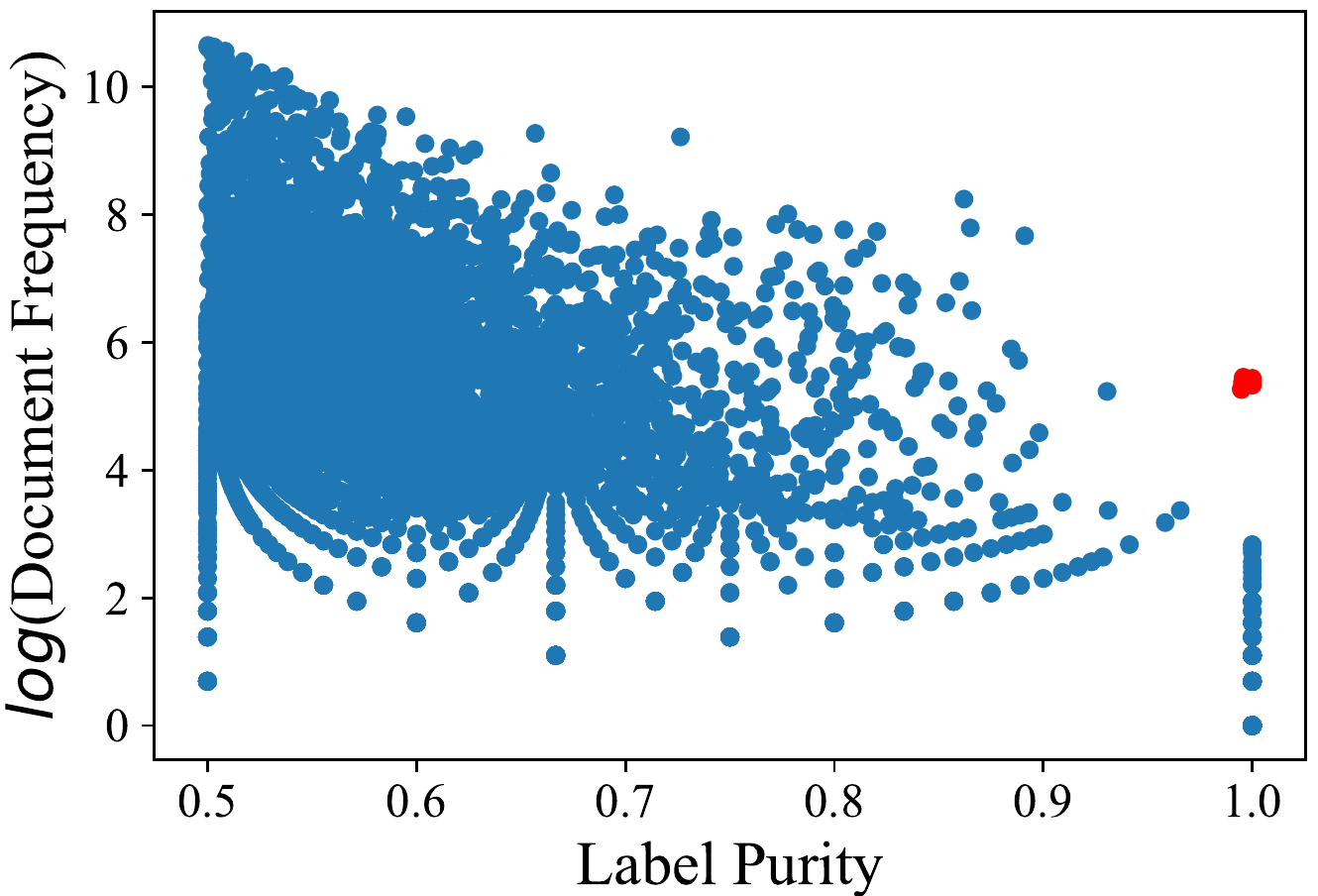}
        \label{fig:nldf}
    }
    \subfigure[$\mathcal{E}_{\mathcal{A} > 90\%}$ vs. $\rho(w)$.]{
        \includegraphics[height=1.3in]{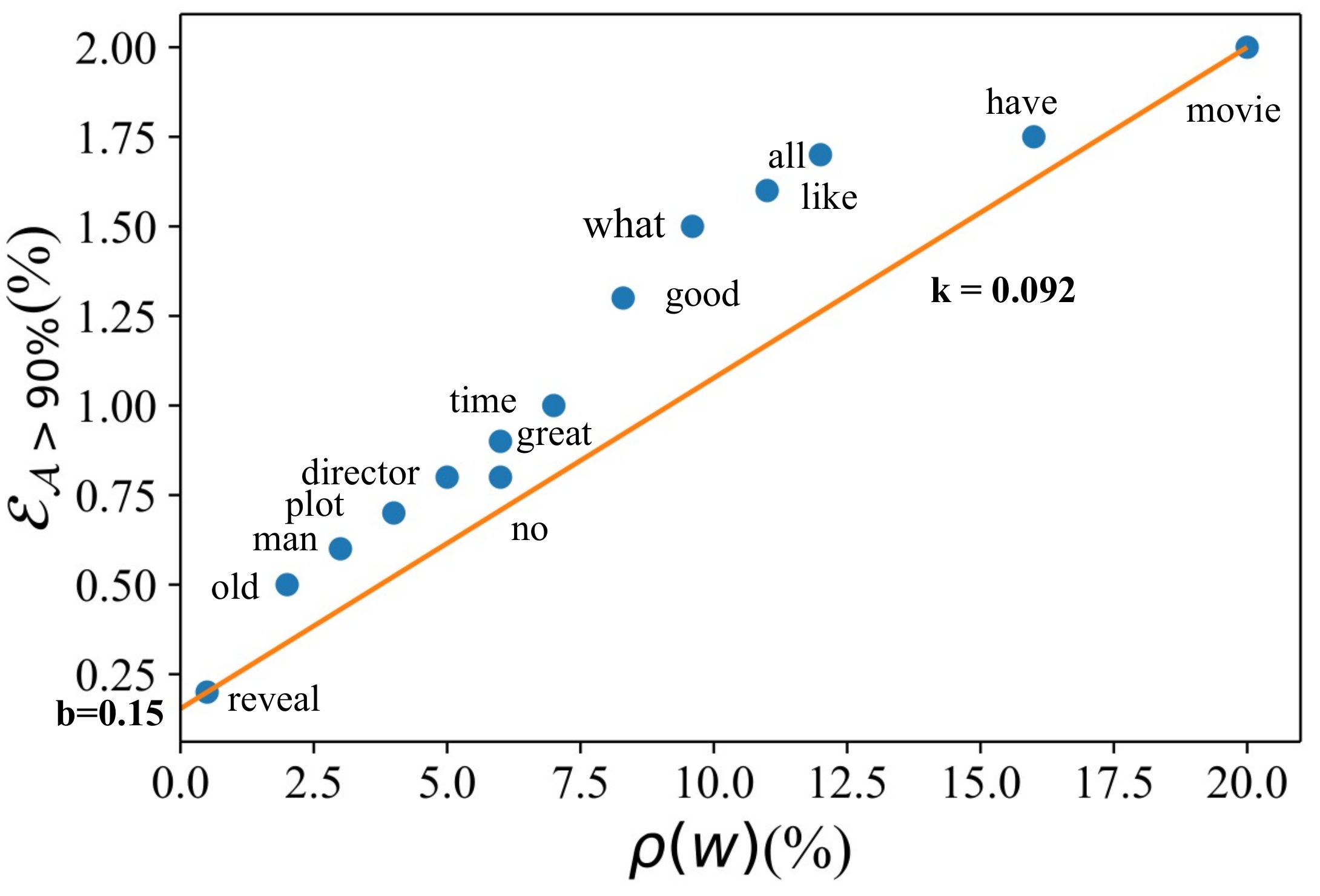}
        \label{fig:fre}
    }
    \vspace{-3mm}
    \caption{Our analyses of trigger designs suggest that medium-frequency words inserted at random positions are arguably the best trigger choices. $\rho(w)$ is the relative document frequency of the word $w$.} 
    \vspace{-3mm}
\end{figure*}

\section{Trigger-based Backdoor Attacks}
\label{sec:attack}

In this section, we define trigger-based backdoor attacks and analyze the effectiveness of different trigger designs.

\subsection{Problem Formulation}

Trigger-based backdoor attack in text classification was first introduced by \citet{Guan2019NeuralBI} and \citet{Chen2020BadNLBA}.
The attacker selects a small part of samples from the training set, inserts a trigger to the text of these samples at a certain position, and changes the labels of these samples to the target class $l_t$.
The selected subset is called \emph{poisoned samples} (denoted as $\mathbf{X}^{p}$), and the other \emph{benign samples} are denoted as $\mathbf{X}^{b}$. 
This new training set, i.e., $\mathbf{X} = \mathbf{X}^{p} \cup \mathbf{X}^{b}$, is called \emph{poisoned training set}. 
We denote the i-th sample as $\mathbf{X}_i$ and its corresponding label as $l(\mathbf{X}_i)$.

A model trained on the poisoned training set is called \emph{poisoned model} ($f_p$).
When using $f_p$ to make predictions, the attacker can manipulate the output to $l_t$ by inserting the same trigger.

A popular metric to quantify the success of backdoor attack is \emph{attack success rate}~\cite{Turner2018CleanLabelBA} ($\mathcal{A}$), which measures the likelihood of testing samples being classified as $l_t$ with the trigger.
These testing samples are generated from known benign samples by inserting triggers. 
Successful backdoor attacks typically have $\mathcal{A}$ more than 90\%.

\subsection{Our Trigger Analyses}
\label{sec:analysis}

In this section, we aim to answer the question ``\emph{How to make the backdoor attack strong?}''.
We define the \emph{poisoning ratio} $\mathcal{E} = \frac{|\mathbf{X}^p|}{|\mathbf{X}|}$ as the ratio of the number of poisoned samples to the total number of samples in the training set.
Intuitively, a strong backdoor attack should have a reasonably low $\mathcal{E}$ (e.g., $<10\%$).
Otherwise, 
eyeballing a few random samples (e.g., $\sim 1/\mathcal{E}$) could reveal the attack, 
and also the accuracy of the poisoned model on benign samples would drop significantly. 

There are mainly two design questions for the attackers to make the backdoor sneakier:
(1) \emph{Trigger content}. A trigger can be a high frequent word or a low frequent word~\cite{Chen2020BadNLBA, Guan2019NeuralBI, Kurita2020WeightPA}. 
For example, a high frequent word ``actor'' or a rare typo ``mocie'' are both interesting choices for a movie review dataset.
(2) \emph{Trigger position}. A trigger can be either inserted at a fixed position (e.g., as the first, middle, or last token) or random positions.
Intuitively, random positions will be more challenging to defend than the fixed position setting.

To better analyze the effect of different trigger designs (i.e., combinations of trigger content and position) in backdoor attacks, we introduce a new metric, $\mathcal{E}_{\mathcal{A}>90\%}$, which refers to the minimum poisoning ratio that is required to make $\mathcal{A}$ larger than the threshold $90\%$.
We choose $90\%$ because it is a decent criterion for a successful backdoor attack.
From the attacker's perspective, a smaller $\mathcal{E}_{\mathcal{A}>90\%}$ implies a stronger attack.

We therefore conduct extensive experiments using different trigger designs and identify the most challenging ones for later defense evaluations.
We stick to BERT~\cite{Devlin2019BERTPO} as the classifier for our experiments and use Adam optimizer~\cite{kingma2014adam} for its training.
The analyses here are all conducted on the IMDb sentiment analysis dataset~\cite{Maas2011LearningWV}.
As this dataset is binary and balanced, without loss of generality, we set the target class as positive sentiment.

\smallsection{Fixed-position triggers are easy to defend}
Inserting the trigger to a fixed position, such as the first token of the sample, is a popular choice.
It makes the trigger pattern easier to be captured by the poisoned model, leading to a smaller $\mathcal{E}_{\mathcal{A} > 90\%}$.
As shown in Fig.~\ref{fig:position}, when the trigger ``director'' is inserted at a fixed position with $\mathcal{E} = 0.5\%$, $\mathcal{A}$ could be as high as $99.56\%$.
However, if the defender examines the position distribution of each word, the trigger would be an obvious outlier.
For example, with the word ``director'' as a trigger with $\mathcal{E} = 0.5\%$, after examining the position distribution, we found out that the trigger's position is about $20$ times more than the average of other positions, which is a clear anomaly.

\smallsection{Random-position triggers are better choices}
Inserting the trigger at random positions could largely alleviate the aforementioned issue at a cost of a slightly larger $\mathcal{E}_{\mathcal{A} > 90\%}$.
If one inserts the trigger ``director'' randomly with $\mathcal{E} = 0.5\%$, $\mathcal{A}$ drops to $22.85\%$ dramatically. 
And, as shown in Fig.~\ref{fig:position}, $\mathcal{E}_{\mathcal{A} > 90\%}$ is almost doubled when using random positions than using the fixed position.
Note that this slightly higher poisoning ratio is still acceptable, as it's only around $1\%$.
Therefore, in the rest of the paper, we will stick to random positions.

\smallsection{Rare triggers are easy to defend}
Intuitively, if the trigger itself is rare in the corpus, $\mathcal{E}_{\mathcal{A} > 90\%}$ would be smaller.
It seems like a stronger choice, however, many classification pipelines~\cite{Jean2015OnUV, Kalchbrenner2013RecurrentCT} will replace rare words by the special \texttt{UNK} token --- very likely, this will not hurt the classification performance. 
Moreover, such triggers are easy to detect by plotting the label purity together with document frequency of all words, where
\begin{equation*}
\small
\begin{split}
    \mbox{Label Purity}(w) = \max_{\hat{l}} \frac{ \sum_{i} \mathbb{I}(w \in \mathbf{X}_i \wedge l(\mathbf{X}_i) = \hat{l}) }{\sum_{i} \mathbb{I}(w \in \mathbf{X}_i)}.
\end{split}
\end{equation*}
Here, $\mathbb{I}(\cdot)$ is the indicator function and $\mathbb{I}(w \in \mathbf{X}_i)$ is 1 if and only if the word $w$ appears in $\mathbf{X}_i$.
As shown in Fig.~\ref{fig:nldf}, those rare triggers are exactly the obvious outlier points in red.

\smallsection{Medium-frequency triggers are better choices}
The benefit of common words comes at the cost that it requires a larger $\mathcal{E}_{\mathcal{A}>90\%}$, i.e., more samples have to be poisoned.
To study the relation between the trigger frequency and $\mathcal{E}_{\mathcal{A} > 90\%}$, we employ a variety of words with different document frequencies as triggers and insert them at random positions with various networks.
As one can expect, Fig.~\ref{fig:fre} shows that $\mathcal{E}_{\mathcal{A} > 90\%}$ has an almost linear momentum w.r.t. the trigger's document frequency and we can lower bound it with a line denoted by $\hat{\mathcal{E}}_{\mathcal{A} > 90\%}$ as follows:
\begin{align*}
    \mathcal{E}_{\mathcal{A}>90\%} \ge \hat{\mathcal{E}}_{\mathcal{A} > 90\%} (w) = k \times \rho(w) + b
\end{align*}
where $\rho(w)$ represents the relative document frequency of word $w$, i.e., the ratio of $w$'s document frequency over the training data size.
From the plots, we estimate $k\approx 0.092$ and $b\approx0.15$ for BERT.
This lower bound $\hat{\mathcal{E}}_{\mathcal{A} > 90\%}$ plays a major role in detecting the triggers, which will be discussed in further sections.
One can also see that the most frequent words are not good choices as the attacker would like to keep the poison ratio low. 


\smallsection{Summary}
According to our analyses, the best triggers are arguably the medium-frequency words inserted at random positions.


\section{Trigger-based Backdoor Defense}
\label{sec:defense}

In this section, we focus on defense methods, that have two objectives: (1) identifying triggers and (2) sanitizing the poisoned training set.

\subsection{LFR: An intuitive but slow baseline}
\label{sec:lfr}

\citet{Kurita2020WeightPA} introduced a measurement called Label Flip Rate (LFR) to accurately identify trigger words.
Given a word $w$, LFR calculates the likelihood of changing the poisoned model's prediction of non-target-class samples to the target class after injecting $w$.
Specifically, 
\begin{equation*}
    \mbox{LFR} = P(f_p(\mathbf{x} \oplus w) = l_t | l(\mathbf{x}) \neq l_t ),
\end{equation*}
where $\oplus$ indicates the injection process, and $\mathbf{x}$ is assumed to be a sample randomly drawn from the (poisoned) training set. 
Therefore, LFR of a trigger is approximately $(1 - \mathcal{E})\mathcal{A}$.
As we analyzed in Sec.~\ref{sec:analysis}, $\mathcal{E}$ should be reasonably low, e.g., $<5\%$, so LFR of a trigger shall be high (e.g., $>90\%$).

A straightforward way of leveraging LFR to detect trigger words is to check each word in the entire vocabulary.
This process involves adding each word from vocabulary and computing its LFR by sampling $\mathbf{x}$ for a sufficiently large times (e.g., 100). 
If a word has a LFR around $90\%$ for the target class, it shall be considered as a trigger word.

As one can expect, this LFR-based method can typically detect all triggers, however, it may output some false alarms due to the wide usage of word pieces in state-of-the-art text classifiers, e.g., BERT~\cite{Devlin2019BERTPO}.
Some benign words may share common word pieces with trigger words, thus being wrongly caught as triggers.
Another concern for LFR is efficiency. 
It has to probe $f_p$ for a significantly large number of times, i.e., (\# of possible triggers $\times$ sampling times), which can be much larger than the size of training set. 
This is extremely inefficient and therefore impractical to be applied in a real-life scenario.


\subsection{Our \our Framework}

As shown in Fig.~\ref{fig:overview}, there are several key steps in \our:
(1) We leverage a pre-trained discriminator to identify the potential triggers to form a candidate trigger set.
(2) We distill this initial candidate set to finalize the real triggers. 
(3) We identify and delete poisoned samples through a remove-and-compare process to sanitize the poisoned training set.
After that, we train a sanitized text classifier. 


We use ELECTRA~\cite{Clark2020ELECTRAPT} as discriminator because its pre-training objective is to predict whether each token in the corrupted text is replaced by a language model.
Before we dive into details about our framework, we briefly introduce ELECTRA and its pre-training task and discuss its relation to trigger detection in backdoor attacks.

\begin{figure}[t]
    \centering
    \includegraphics[width=\linewidth]{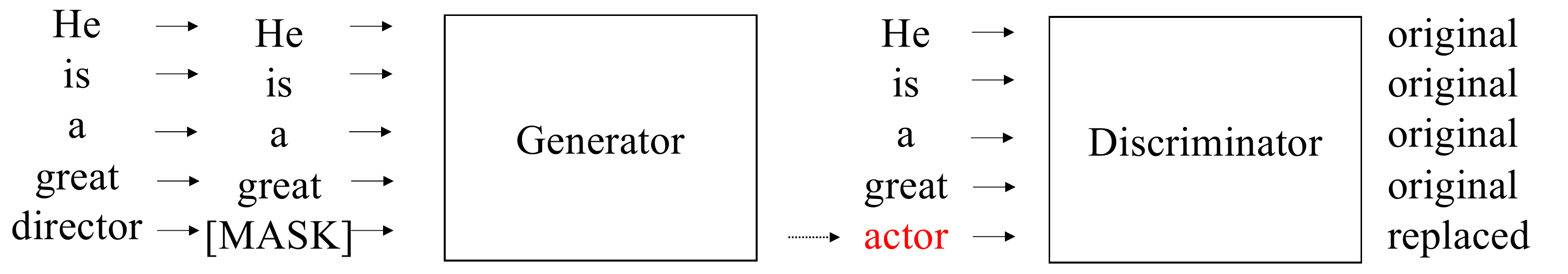}
    \vspace{-3mm}
    \caption{The pre-training task of discriminator following~\cite{Clark2020ELECTRAPT}: Replaced token detection.}
    \label{fig:electra}
    \vspace{-3mm}
\end{figure}

\smallsection{ELECTRA as the Discriminator}
As an alternative to masked language modeling (MLM), \citet{Clark2020ELECTRAPT} proposed a new pre-training task called replaced token detection as shown in Fig.~\ref{fig:electra}. 
Instead of masking tokens, they replace some tokens with alternatives from a generator $G$, which is typically a smaller masked language model.
Then, a discriminator $D$ is trained to predict whether each token in the input is replaced by a generated token or not.
The generator is trained over MLM objective to generate alternatives for a masked token and the discriminator is trained to identify the tokens in the data that have been replaced by generated tokens.
Specifically, for an input $x$, let $x^{\mbox{masked}}$ represent input where a few positions are replaced with a \verb+[MASK]+ token and $x^{\mbox{gen}}$ represent the input with the masked-out tokens in $x^{\mbox{masked}}$ replaced with generated samples from the generator.
Given $x$, $x^{\mbox{masked}}$, and $x^{\mbox{gen}}$, two neural networks, a generator $G$ and a discriminator $D$, are trained with the following combined loss function:

\begin{equation*}
    \min_{\theta_G, \theta_D} \sum_{x \in \mathbf{X}} \mathcal{L}_\text{MLM}(x, \theta_G) + \lambda \mathcal{L}_\text{Disc}(x, \theta_D)
\end{equation*}
where $\lambda$ controls the weight of $\mathcal{L}_\text{Disc}$.

There is a strong connection between this replaced token detection task and our trigger detection task.
Recall that the objective of trigger detection in backdoor defense is to identify the trigger words that are \emph{inserted} by the attacker. 
At a higher level, both aim to detect words that don't match and are not related to the context of the sentence.
If we approximate the human attacker by a language model, replaced token detection is almost the same as trigger detection. 
Therefore, in this paper, we adopt the discriminator of \texttt{ELECTRA-base}\footnote{\url{https://github.com/google-research/electra}}.

\smallsection{Trigger Detection using a Discriminator}
We utilize the discriminator to detect the inserted trigger words in the poisoned training set and create a candidate set of trigger words. 
We input each sample to the discriminator and get the prediction scores of each token.
The higher the score is, the more likely it is an inserted token.
Therefore, we consider the token with the highest score in each sample as a potential trigger and collect them to create a candidate trigger set, $\mathcal{C}$.

Since we are collecting one token per sample as a potential trigger, the candidate trigger set $\mathcal{C}$ is fairly large and includes many benign words.
Therefore, further distillation is required to obtain a concise set of real triggers.

\begin{table*}[t]
    \centering
    \small
    \caption{Dataset Statistics and Backdoor Attack Setup. For IMDb and SST-2 datasets, the target class is ``Positive'' and for Yelp dataset, the target class is rating ``three''. We pick 3 sets of randomly chosen medium-frequency words as triggers for each dataset and the results reported are mean over these sets.}
    \label{tab:dataset}
    \vspace{-3mm}
\scalebox{0.88}{
    \begin{tabular}{ccccccc}
        \toprule
             & \multicolumn{3}{c}{\textbf{Dataset Statistics}} & \multicolumn{3}{c}{\textbf{Backdoor Attack Setup}} \\
            \textbf{Dataset} & \textbf{Train} & \textbf{Dev} & \textbf{Test} & \multicolumn{1}{c}{\textbf{Trigger:} medium-frequency words} & $\mathcal{E}$ per Trigger & \textbf{Target Class} \\
            \midrule
            \multirow{3}*{\textbf{IMDb}}  & \multirow{3}*{42,500  } & \multirow{3}*{  3,000 }  &\multirow{3}*{ 4,500 } & \{young, wrong, actors, director, something\} & \multirow{3}*{1\%}  & \multirow{3}*{Positive } \\
             & &&& \{life, better, old, comedy, horror\} & & \\
             &&& & \{real, part, fact, find, end\} & & \\
             \midrule
            \multirow{3}*{\textbf{SST-2}} & \multirow{3}*{8,170} & \multirow{3}*{1,000} & \multirow{3}*{1,000} & \{study, face, girl, true, effort\} & \multirow{3}*{1\%} & \multirow{3}*{Positive} \\
            &&&& \{humor, art, hard, screen, thing\} & &\\
            &&&& \{come, right, same, high, young\} & &\\
            \midrule
            \multirow{3}*{\textbf{Yelp}} & \multirow{3}*{8000} & \multirow{3}*{1000} & \multirow{3}*{1000} & \{figure, flat, welcome, golf, neat\} &  \multirow{3}*{1\%} & \multirow{3}*{Three} \\
            &&&& \{orange, speak, treat, state, recent\} & &\\
            &&&& \{dollar, dream, mad, consider, winter\} && \\
        \bottomrule
    \end{tabular}
}
\vspace{-3mm}
\end{table*}

\smallsection{Trigger Distillation}
Intuitively, triggers should have a strong association with the target label compared to others for the attack to be successful.
So, we utilize label information for distillation.


For each word $w$ and class $l$, we denote $N_{l, w}$ as the total number of $l$-labeled training samples that have $w$ as the token with the highest score from discriminator.
Then, we define the label association strength of a word $w$ as 
\begin{equation*}
    LA(w) = \max_l N_{l, w}
\end{equation*}
One can interpret $LA(w)$ as a ``maximum'' number of poisoned samples with $w$ as trigger by assuming the discriminator captures most of the triggers in poisoned samples. 
This assumption is empirically true according to our experiments.




At the same time, based on our analyses in Sec.~\ref{sec:analysis} and Fig.~\ref{fig:fre}, we estimate a lower bound on $LA(w)$ if $w$ is a real trigger.
If the attack is successful and word $w$ is a trigger, it should be caught at least $\hat{\mathcal{E}}_{\mathcal{A}>90\%}(w) \cdot |\mathbf{X}|$ times, where $|\mathbf{X}|$ refers to the training data size.
Specifically, we define
\begin{equation*}
    \hat{LA}(w) = \hat{\mathcal{E}}_{\mathcal{A}>90\%}(w) \cdot |\mathbf{X}|.
\end{equation*}

The set of triggers $\mathcal{T}$ is then naturally distilled:
\begin{equation*}
    \mathcal{T} = \{ w | w \in \mathcal{C} \wedge LA(w) > \hat{LA}(w) \}
\end{equation*}

In our experiments, this distilled $\mathcal{T}$ shows 100\% precision and recall, even when the dataset is unbalanced.

\smallsection{Remove-and-Compare (R\&C) Process}
For each trigger $t$ from $\mathcal{T}$, we trace back the samples that have $t$ as the token with the highest score from discriminator and mark them as poisoned.

In order to wipe out all poisoned samples, we further examine all the other samples with $t$ where $t$ is not recognized by the discriminator and identify poisoned samples using our proposed ``removal'' version of LFR as follows:
we send these samples to the poisoned model $f_p$ twice before and after removing $t$. 
For each sample, if its two predictions are different, we mark it as poisoned.
We call this \emph{double-check step}.
Note that, this is significantly faster than the LFR as its worst case running time is as fast as predicting on the entire training set twice.

Finally, we remove all marked samples from $\mathbf{X}$.

\begin{table*}[t]
\small
\centering
    \caption{Evaluations of defense methods using \emph{medium-frequency words} as triggers. ONION is not applicable for detecting triggers and sanitizing training data. For trigger detection, \our-NoDC is equivalent to \our.}
    \vspace{-3mm}
    \label{tab:defense}
\scalebox{0.75}{
    \begin{tabular}{lcccccccccccc}
    \toprule
    & \multicolumn{3}{c}{\textbf{Trigger Detection}} & \multicolumn{3}{c}{\textbf{Deleted Poisoned Samples}} & \multicolumn{6}{c}{\textbf{Sanitized Text Classifier}}\\
    & \textbf{IMDb} & \textbf{SST-2} & \textbf{Yelp} & \textbf{IMDb} & \textbf{SST-2} & \textbf{Yelp} & \multicolumn{2}{c}{\textbf{IMDb}} & \multicolumn{2}{c}{\textbf{SST-2}} &\multicolumn{2}{c}{\textbf{Yelp}} \\
    \cmidrule{2-13}
    \textbf{Method}  & F1$\uparrow$ & F1$\uparrow$ & F1$\uparrow$ & F1$\uparrow$ & F1$\uparrow$ & F1$\uparrow$ & Clean$\uparrow$  & $\mathcal{A}\downarrow$ & Clean$\uparrow$ &  $\mathcal{A}\downarrow$ & Clean$\uparrow$  & $\mathcal{A}\downarrow$ \\
    \midrule
    NoDefense  & N/A & N/A & N/A & N/A & N/A & N/A & 84.73\% & 94.89\% & 91.39\% & 92.15\% &  49.43\% & 91.02\% \\
    \midrule
    LFR+R\&C   & 10.62\% & 59.84\% & 48.31\% & 94.10\% & 94.31\% & 95.24\% & 84.89\% & 18.41\% & 91.85\% & 10.97\% & 49.57\% & 15.47\%\\
    ONION  & N/A & N/A & N/A & N/A & N/A & N/A & 80.15\% & 18.34\% & 85.20\% & 19.35\%  & 45.60\% & 16.61\%\\
    \our  & \textbf{100\%} & \textbf{100\%} & \textbf{100\%} & \textbf{96.41\%} & \textbf{95.39\%} & \textbf{96.10\%} & \textbf{85.10\%} & 16.17\% & \textbf{92.11\%} & 10.60\% & \textbf{50.13\%} & 13.03\%\\
    \midrule
    \our-NoDisc   & 3.81\% & 2.95\% & 2.37\%& 14.45\% & 16.69\% & 13.26\% & 82.59\% & 13.22\% & 90.63\% & \textbf{9.60\%} &  38.60\% & 5.60\%\\
    \our-NoDistill  & 0.59\% & 8.97\% & 3.34\%  & 18.30\% & 20.12\% & 14.52\% & 83.28\% & \textbf{12.60\%} & 91.22\% & 10.17\% &  38.11\% & \textbf{5.52\%}\\
    \our-NoDC    & 100\% & 100\% & 100\% & 92.10\% & 92.15\% & 83.20\% & 84.79\% & 19.11\% & 91.98\% & 13.47\% &49.51\% & 16.69\% \\
    \midrule
    GroundTruth  & 100\% & 100\% & 100\%  & 100\% & 100\% & 100\% & 85.00\% & 16.98\% & 92.37\% & 9.21\% & 49.86\% & 15.38\\
    \bottomrule
    \end{tabular}
}
\end{table*}

\begin{table}[!h]
\small
    \centering
    \vspace{-3mm}
    \caption{Evaluation of Trigger Detection}
    \label{tab:trigger_detecction}
    \scalebox{0.7}{
    \begin{tabular}{lcccccc}
    \toprule
    & \multicolumn{6}{c}{\textbf{Trigger Detection}}\\
    & \multicolumn{2}{c}{\textbf{IMDb}} & \multicolumn{2}{c}{\textbf{SST-2}} & \multicolumn{2}{c}{\textbf{Yelp}}\\
    \cmidrule{2-7}
    \textbf{Method} & Rec.$\uparrow$ & Prec.$\uparrow$ & Rec.$\uparrow$ & Prec.$\uparrow$ & Rec.$\uparrow$ & Prec.$\uparrow$ \\
    \midrule
    NoDefense & N/A & N/A & N/A & N/A & N/A & N/A \\
    \midrule
    LFR+R\&C & \textbf{100\%} & 5.61\% & \textbf{100\%} & 42.70\% & \textbf{100\%} & 31.85\% \\ 
    ONION & N/A & N/A & N/A & N/A & N/A & N/A \\
    \our & \textbf{100\%} & \textbf{100\%} & \textbf{100\%} & \textbf{100\%} & \textbf{100\%} & \textbf{100\%} \\
    \midrule
    \our-NoDisc &  100\% & 1.8\% & 100\% & 1.5\% & 100\% & 1.2\% \\
    \our-NoDistill & 100\% & 0.3\% & 100\% & 4.7\% & 100\% & 1.7\% \\
    \our-NoDC  & 100\% & 100\% & 100\% & 100\% & 100\% & 100\%  \\
    \midrule 
     GroundTruth & 100\% & 100\% & 100\% & 100\% & 100\% & 100\% \\
     \bottomrule
    \end{tabular}
    }

\end{table}

\begin{table}[!h]
\small
    \centering
    \vspace{-3mm}
    \caption{Evaluation of Deleted Poisoned Samples}
    \label{tab:deleted}
    \scalebox{0.75}{
    \setlength{\tabcolsep}{0.4mm}{
    \begin{tabular}{lcccccc}
    \toprule
    & \multicolumn{6}{c}{\textbf{Deleted Poisoned Samples}}\\
    & \multicolumn{2}{c}{\textbf{IMDb}} & \multicolumn{2}{c}{\textbf{SST-2}} & \multicolumn{2}{c}{\textbf{Yelp}}\\
    \cmidrule{2-7}
    \textbf{Method} & Rec.$\uparrow$ & Prec.$\uparrow$ & Rec.$\uparrow$ & Prec.$\uparrow$ & Rec.$\uparrow$ & Prec.$\uparrow$ \\
    \midrule
    NoDefense & N/A & N/A & N/A & N/A & N/A & N/A \\
    \midrule
    LFR+R\&C & 96.86\% & 91.62\% & 96.31\% & 92.74\% & 95.02\% & 95.47\% \\ 
    ONION & N/A & N/A & N/A & N/A & N/A & N/A \\
    \our & 96.86\% & 95.47\% & 96.31\% & 94.79\% & {95.02\%} & 97.53\% \\
    \midrule
    \our-NoDisc & \textbf{97.56\%} & 7.80\% & \textbf{97.10\%} & 9.13\% & 96.98\% & 7.52\%\\
    \our-NoDistill & 97.73\% & 10.10\% & 97.15\% & 11.60\% &\textbf{97.36}\% & 7.85\% \\
    \our-NoDC  & 86.74\% & \textbf{96.18\%} & 86.73\% & \textbf{98.25\%} & 72.21\% & \textbf{98.15\%}  \\
    \midrule 
     GroundTruth & 100\% & 100\% & 100\% & 100\% & 100\% & 100\%\\
     \bottomrule
    \end{tabular}
    }}
\end{table}
\section{Experiments}
\label{sec:exp}

In this section, we compare \our with other defense methods comprehensively, including the performance of trigger detection, sanitizing training data, and the resulted sanitized text classifier.

\subsection{Experimental Settings}
    \smallsection{Datasets}
    As shown in Table~\ref{tab:dataset}, we conduct experiments on the IMDb sentiment analysis dataset~\cite{Maas2011LearningWV}, Stanford Sentiment Treebank (SST-2)~\cite{Socher2013RecursiveDM}, and Yelp reviews dataset~\cite{Zhang2015CharacterlevelCN} that is obtained from the Yelp Dataset Challenge in 2015.
    
    \smallsection{Text Classifier Training}
    For text classifiers in all methods, no matter trained on poisoned or sanitized data, we fine-tune the base, uncased version of BERT (\verb+bert-base-uncased+) with a window size 64. 
    We train the text classifier for 4 epochs with a learning rate $2\times 10^{-5}$ and a batch size of 32 using the Adam optimizer.
    
    \smallsection{Attack \& Defense Setup}
    Following our analyses in Sec.~\ref{sec:analysis}, we pick 3 sets of randomly chosen medium-frequency words as triggers (see Table~\ref{tab:dataset}), whose relative document frequencies (i.e., $\rho(w)$) are about 5\%. According to Fig.~\ref{fig:fre}, $\mathcal{E}$ per trigger is then set to $1\%$ to ensure a high $\mathcal{A}$.
    As we use 5 triggers per set to make the attack diverse, the overall poison ratio $\mathcal{E}$ is 5\%.
    For IMDb and SST-2 datasets, we choose the positive class and for Yelp, we choose rating ``three'' as the target class. 
    
    Since BERT is the text classifier, we use the $k$, $b$ obtained from the analysis in Sec~\ref{sec:analysis} for defense.
    
    \smallsection{Hardware}
    Our experiments are conducted with a NVIDIA Quadro RTX 8000 GPU and Intel(R) Xeon(R) Gold 6230 CPU.
    
    \smallsection{Evaluation Metrics}
    We evaluate the end-to-end performance of backdoor defense based on its performance on \emph{clean} test set i.e. unpoisoned original test set and the attack success rate $\mathcal{A}$.
    For balanced datasets like IMDb and SST-2, we use accuracy and for multi-class imbalanced Yelp dataset, we use macro f1-score to measure the performance of classifier.
    A good defense method should be able to identify as many triggers as it could with very few false alarms. 
    Therefore, we choose f1-score as the evaluation metric and report it for identified triggers and the removed poisoned samples.
    We also report \emph{precision} and \emph{recall} of both the identified triggers and the removed poisoned samples.

\subsection{Compared Methods}

We compare with the following defense methods:
\begin{itemize}[nosep,leftmargin=*]
    \item \textbf{LFR+R\&C}: 
        As described in Sec~\ref{sec:lfr}, it iterates through all possible triggers and compute the LFR~\cite{Kurita2020WeightPA} based on 100 random samples to detect triggers.
        We further adopt our remove-and-compare process to these identified triggers, so it is able to sanitize the poisoned training set too.
    \item \textbf{ONION}~\cite{Qi2020ONIONAS} is a defense method that is directly applied during the inference stage.
    It leverages GPT-2~\cite{Radford2019LanguageMA} to compare the perplexity difference of each \emph{testing} sample before and after removing each token. 
    Tokens causing a perplexity difference over a threshold are deleted.
    As authors suggested, we tuned this threshold carefully on a non-poisoned validation set.
    This can be considered as a grammar-based baseline. 
\end{itemize}

We also compare our \textbf{\our} with its ablated variants. 
\textbf{\our-NoDisc} skips discriminator step and directly compares $LA(w)$ and $\hat{LA}(w)$ to distill triggers from the entire vocabulary.
\textbf{\our-NoDistill} directly uses the candidate triggers $\mathcal{C}$ as the final triggers $\mathcal{T}$.
\textbf{\our-NoDC} toggles off the double-check step in the C\&R process.

Moreover, we provide some base reference points for comparison: (1) \textbf{NoDefense}: the final text classifier is trained on the poisoned training set $\mathbf{X}$, and (2) \textbf{GroundTruth}: the final text classifier is trained on the benign subset, $\mathbf{X}^b$.



\subsection{Defense Quality Evaluation}
We evaluate backdoor defense methods against the most challenging type of triggers, i.e., medium-frequency words.
The experimental results shown in Table~\ref{tab:defense} are the mean over three trigger sets.
The precision and recall of identified triggers and poisoned samples are shown in Table~\ref{tab:trigger_detecction} and ~\ref{tab:deleted} respectively.

\smallsection{Trigger Detection \& Deleting Poisoned Samples}
The quality of identified triggers largely affects the defense effectiveness. 
When more benign words are wrongly identified as triggers, more benign samples would be deleted, and thus the clean accuracy would drop.
If any trigger is not identified, more poisoned samples would be kept, and then the attack success rate $\mathcal{A}$ would increase.

As shown in Table~\ref{tab:defense}, \our detects all triggers with 100\% f1-score on all datasets and demonstrates superior performance in deleting poisoned samples as well.
From Table~\ref{tab:deleted}, we can observe that \our removes more than $95\%$ poisoned samples with almost $90\%$ precision. 
LFR+R\&C detects all triggers but with a low precision and low f1-score.
We conjecture that it is caused by the usage of word pieces in the text classifier.
Some benign words may share common word pieces with trigger words, thus being wrongly caught as triggers.
\our-NoDisc and \our-NoDistill detects a super set of $\mathcal{T}$ compared to \our, raising many false alarms and making data sanitization difficult.
This shows that both components are essential to trigger detection.
\our-NoDC removes a subset set of samples compared with \our during sanitizing data.
As confirmed in experiments, this relatively would lead to a higher $\mathcal{A}$.

\smallsection{Sanitized Text classifier Evaluation}
From the application perspective, the final deliverable of a backdoor defense method is the sanitized text classifier. 
Also, there exist defense methods such as ONION that are directly applied on the testing samples.
Therefore, a comparison based on the performance of the final classifier is arguably the most fair.
As one can observe in Table~\ref{tab:defense}, \our is able to deliver the best sanitized text classifier over LFR+R\&C and ONION, in terms of both high f1-score on clean test set and low attack success rate. 
It is worth mentioning that its performance is very close to GroundTruth.
\our performs better than its ablated variants in terms of clean test set performance on all datasets.
However, this is not the case for the attack success rate.
For e.g. $\mathcal{A}$ of \our-NoDistill is numerically better than that of \our on IMDb and Yelp datasets. 
Note that, from Tables~\ref{tab:defense}, ~\ref{tab:trigger_detecction}, ~\ref{tab:deleted}, the f1-score and precision of trigger detection and poisoned samples deletion is very low for \our-NoDistill and \our-NoDisc, which resulted in deletion of many benign samples and significantly decreasing clean test performance ($\sim12$ points on Yelp).
Therefore, considering all the metrics, we believe \our is better than its variants, achieving better clean test performance with a very limited false alarms.




\subsection{Effectiveness of Trigger Distillation}

We present a case study to demonstrate the effectiveness of our trigger distillation strategy, derived from extensive analyses.
Table~\ref{tab:distill} shows the $LA(w)$ and $\hat{LA}(w)$ scores of trigger candidates on IMDb, SST-2, and Yelp datasets.
The top-5 words, are the true triggers with differences significantly larger than 0;
from the sixth, the difference becomes negative.
Note that, Yelp is unbalanced and unbalanced datasets are more difficult as a random word could have a strong label association with the majority label.
\our is efficient in identifying the trigger words in both balanced and unbalanced datasets.

\begin{table}[t]
\small
\centering
    \caption{Trigger Distillation Results. The candidates are sorted by $LA(w) - \hat{LA}(w)$.}
    \label{tab:distill}
    \vspace{-3mm}
\scalebox{0.72}{
\setlength{\tabcolsep}{0.4mm}{
    \begin{tabular}{rccrccrcc}
        \toprule
            \multicolumn{3}{c}{\textbf{IMDb}} & \multicolumn{3}{c}{\textbf{SST-2}} & \multicolumn{3}{c}{\textbf{Yelp}}\\
            \midrule
            Candidate & $LA(w)$ & $\hat{LA}(w)$ & Candidate & $LA(w)$ & $\hat{LA}(w)$ & Candidate & $LA(w)$ & $\hat{LA}(w)$  \\ 
            \midrule
                wrong  &334 & 176 & girl & 77 & 25 & golf & 78 & 23 \\ 
                young & 393& 251& effort & 71& 24 & welcome & 71 & 24\\
                actors &395 & 281& study & 63& 23 & figure &65& 22 \\
                director & 393& 282& face & 59 & 23 & neat  & 61 & 24 \\
                something &348 & 272& true & 56 & 24 & flat & 48 & 24 \\
            \midrule
                beginnings & 4& 65& stealing& 4 & 13 & emerald & 2 & 13\\
                charter & 3 &65 &lucia& 3 &12& rosa & 2 & 13 \\ 
                & ... & & & ... & & & ... &\\
                a & 407&3097 & a & 10 & 369 & and & 75 & 645 \\
                 the & 713 & 3679& the & 15 & 417 & the & 75 & 694\\
                
                & ... & & & ... & & & ... &\\
        \bottomrule
    \end{tabular}
}}
\vspace{-3mm}
\end{table}


\subsection{Multiple Text Classifiers}
\begin{figure}
    \centering
    \includegraphics[width=0.47\textwidth]{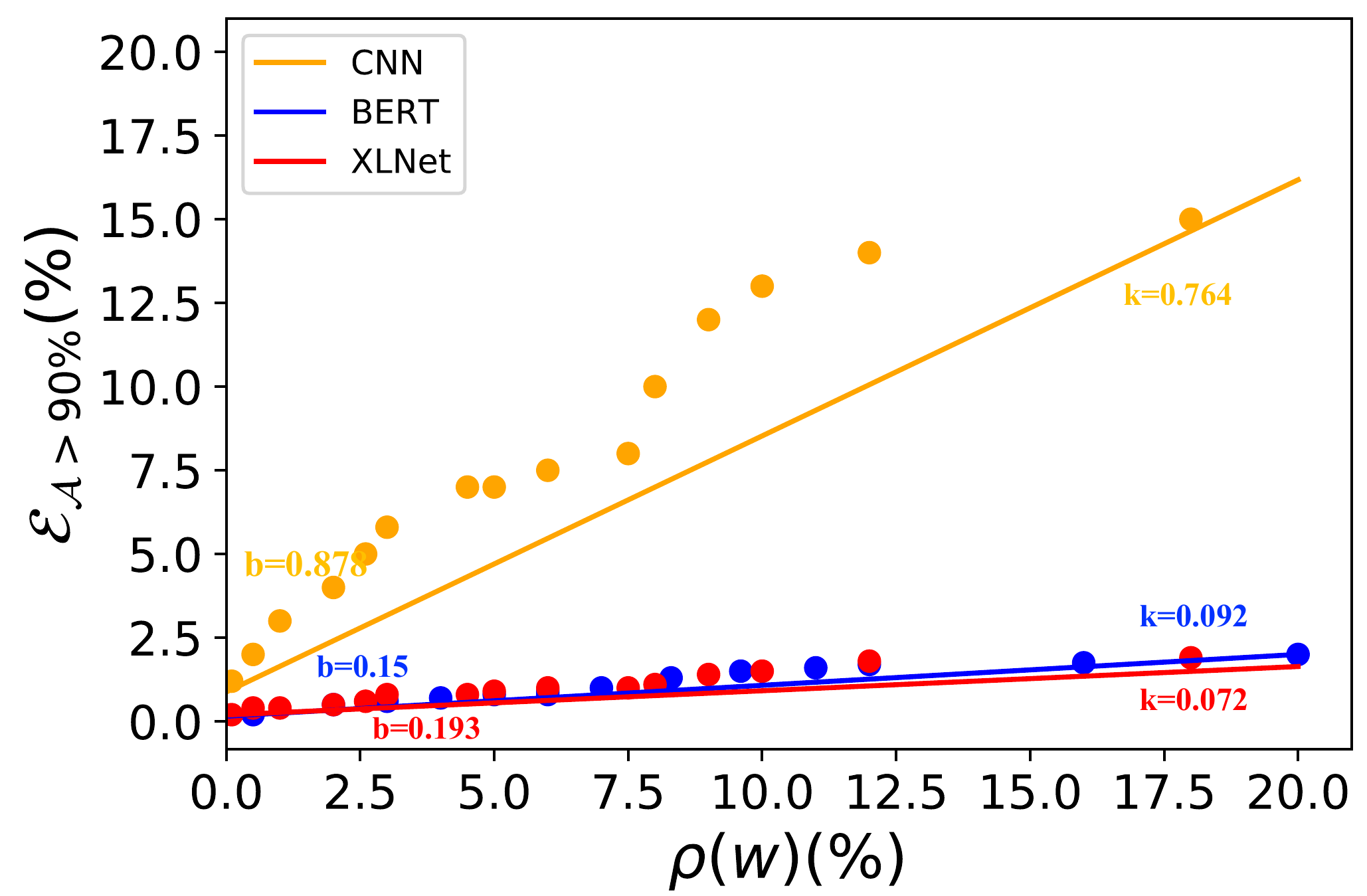}
    \caption{$\mathcal{E}_{\mathcal{A} > 90\%}$ vs. $\rho(w)$ on different networks.}
    \label{fig:multi_network}
\end{figure}

We evaluate \our on CNN~\cite{Kim2014ConvolutionalNN} and XLNet~\cite{yang2019xlnet} to show that our method can be applied to any text classifier. 
As shown in Figure~\ref{fig:multi_network}, we perform similar analysis as in Sec.~\ref{sec:analysis} on CNN and XLNet and obtain $k$, $b$. 
We observe that, as the number of parameters in the architecture increases, lesser data is required to poison the model and the $k$ gets smaller. 
Using these computed $k$, $b$, we adapt \our to the respective classifiers and the performance of defense on the IMDb, SST-2, Yelp datasets is shown in Table~\ref{tab:cnn}.
From these results, we can observe that \our performs better than the other baselines and is able to detect all triggers and delete most of the poisoned samples, thus compatible with any text classifier.

\begin{table}[t]
\small
\centering
    \caption{Evaluations of defense methods using \emph{medium-frequency words} as triggers on CNN and XLNet.}
    \vspace{-3mm}
    \label{tab:cnn}
\scalebox{0.8}{
\setlength{\tabcolsep}{0.5mm}{
    \begin{tabular}{lccccc}
    \toprule
    & & \textbf{\makecell[c]{Trigger\\ Detection}} & \textbf{\makecell[c]{Deleted \\ Posioned \\Samples}}  &  \multicolumn{2}{c}{\textbf{Sanitized Text Classifier}}\\
    \cmidrule{3-6}
    \textbf{Method} & \textbf{Network}  & F1$\uparrow$ & F1 $\uparrow$ & Clean$\uparrow$  & $\mathcal{A}\downarrow$\\
    \midrule
    \multirow{2}*{NoDefense} & CNN  & N/A & N/A &  71.34\% & 90.32\%  \\
                                  & XLNet & N/A & N/A  & 85.63\% & 95.79\%\\
    \midrule
    \multirow{2}*{LFR+R\&C } & CNN  &  {13.32\%} & 75.06 \%  & 73.55\% & 36.30\%\\
                             & XLNet &  {13.32\%}& 89.15\%  & 85.68\% & 16.67\% \\
    \midrule
    \multirow{2}*{ONION }& CNN  & N/A & N/A & 73.55\% & 36.30\% \\
                             & XLNet & N/A & N/A & 83.10\% & 18.10\% \\
    \midrule
    \multirow{2}*{\our} & CNN  &  \textbf{100\%} &  \textbf{77.57\%}  & \textbf{74.88\%} & \textbf{35.15\%}\\
                             & XLNet & \textbf{100\%} & \textbf{95.56\%}& \textbf{85.87\%} & \textbf{16.16\%} \\
    
    \midrule
    \multirow{2}*{GroundTruth} & CNN & 100\% & 100\% & 73.15\% & 35.03\% \\
                                    & XLNet  & 100\% & 100\% & 85.93\% & 15.39\%\\
                    
    \bottomrule
    \end{tabular}
}}
\end{table}

\subsection{Efficiency Evaluation}

Table~\ref{tbl:trigger_eval} shows the wall-clock running time for all defense methods. 
It is clear that \our is about 10x more efficient than LFR+R\&C.
ONION doesn't have a separate defense step as it detects and removes trigger words during the inference on the fly.
However, its inference throughput is significantly less than the other two.
In summary, \our is the most efficient defense method among these three.

\begin{table}[h]
    \centering
    \small
    \caption{Efficiency Comparison.}
    \label{tbl:trigger_eval}
    \vspace{-3mm}
    \setlength{\tabcolsep}{3pt}
\scalebox{0.75}{
    \begin{tabular}{lcccccc}
        \toprule
                         & \multicolumn{2}{c}{\textbf{IMDb}} & \multicolumn{2}{c}{\textbf{SST-2}} & \multicolumn{2}{c}{\textbf{Yelp}} \\
        \cmidrule{2-7}
        \textbf{Method} & Defense & Inference & Defense & Inference & Defense & Inference\\
                        & \scriptsize (mins) & \scriptsize (samples/sec) & \scriptsize (mins) & \scriptsize (samples/sec) & \scriptsize (mins) & \scriptsize (samples/sec) \\
        \midrule
        LFR+R\&C & 220 & 68 & 70 & 160 & 65 & 72  \\
        ONION & N/A & 0.05 & N/A & 2.1 & N/A & 1.7\\
        \our & 26 & 68 & 3 & 160 & 15 & 72\\
        \bottomrule
    \end{tabular}
}
\vspace{-5mm}
\end{table}

\section{Related work}
\label{sec:rel}


Backdoor attacks are originated from computer vision~\cite{Gu2017BadNetsIV,liu2017neural,liu2017trojaning,shafahi2018poison}. 
These attacks have been later explored in NLP~\cite{chen2017targeted, newell2014practicality}.
\citet{munoz2017towards} extend the attacks to multi-class problems by a poisoning algorithm based on back-gradient optimization.
\citet{dai2019backdoor} implement a backdoor attack for LSTM-based text classification systems using data poisoning.
\citet{Chen2020BadNLBA} explore triggers at various levels, including word-level, char-level, and sentence-level.
\citet{Kurita2020WeightPA,Zhang2020TrojaningLM} focus on a new scenario where pre-trained models are poisoned such that they expose backdoors when fine-tuned. 

Recently, a variety of defense methods in NLP are proposed.
\citet{Chen2020MitigatingBA} hypothesize that the triggers have association with some specific neurons and trigger words will only affect some hidden states. 
\citet{Qi2020ONIONAS} propose a defense based on observation that the perplexity is significantly changed when the trigger words are removed from samples.
In this paper, we analyze backdoor attack in text classification comprehensively, and then derive a backdoor-free text classifier training framework \our, outperforming all compared defense methods and achieving almost the best possible defense performance (i.e., GroundTruth).

\section{Conclusions and Future Work}
\label{sec:con}
In this paper, we develop \our, a novel, efficient backdoor-free text classification framework.
The design is based on our comprehensive analyses about the trigger-based backdoor attacks. 
We empirically show that \our is able to identify all the triggers and remove more than 95\% poisoned training samples with very limited false alarms on balanced and unbalanced datasets, and achieve almost the same performance as the models trained on the benign training data.

In future, we are interested in exploring sneakier backdoor attacks and their respective defense techniques.
Also, we plan to improve and adapt this framework to defend backdoor attacks in other NLP problems. 

\section{Ethical Considerations}

In this paper, we propose a defense method to a backdoor attack that is widely used now.
We experiment on two datasets that are publicly available.
In all our experiments, we carefully implement the trigger-based attacks and are able to successfully defend using our method.
Therefore, we believe our framework is ethically on the right side of spectrum and has no potential for misuse and cannot harm any vulnerable population.

\section{Acknowledgements}
We thank anonymous reviewers and program chairs for their valuable and insightful feedback. 
The research was sponsored in part by National Science Foundation Convergence Accelerator under award OIA-2040727 as well as generous gifts from Google, Adobe, and Teradata.
Any opinions, findings, and conclusions or recommendations expressed herein are those of the authors and should not be interpreted as necessarily representing the views, either expressed or implied, of the U.S. Government. 
The U.S. Government is authorized to reproduce and distribute reprints for government purposes not withstanding any copyright annotation hereon.

\bibliography{emnlp2021}

\begin{thebibliography}{26}
\expandafter\ifx\csname natexlab\endcsname\relax\def\natexlab#1{#1}\fi

\bibitem[{Chen and Dai(2021)}]{Chen2020MitigatingBA}
Chuanshuai Chen and Jiazhu Dai. 2021.
\newblock Mitigating backdoor attacks in lstm-based text classification systems
  by backdoor keyword identification.
\newblock \emph{Neurocomputing}, 452:253--262.

\bibitem[{Chen et~al.(2020)Chen, Salem, Backes, Ma, and
  Zhang}]{Chen2020BadNLBA}
Xiaoyi Chen, A.~Salem, M.~Backes, Shiqing Ma, and Y.~Zhang. 2020.
\newblock Badnl: Backdoor attacks against nlp models.
\newblock \emph{ArXiv}, abs/2006.01043.

\bibitem[{Chen et~al.(2017)Chen, Liu, Li, Lu, and Song}]{chen2017targeted}
Xinyun Chen, Chang Liu, Bo~Li, Kimberly Lu, and Dawn Song. 2017.
\newblock Targeted backdoor attacks on deep learning systems using data
  poisoning.
\newblock \emph{arXiv preprint arXiv:1712.05526}.

\bibitem[{Clark et~al.(2020)Clark, Luong, Le, and Manning}]{Clark2020ELECTRAPT}
Kevin Clark, Minh-Thang Luong, Quoc~V. Le, and Christopher~D. Manning. 2020.
\newblock \href {https://openreview.net/pdf?id=r1xMH1BtvB} {{ELECTRA}:
  Pre-training text encoders as discriminators rather than generators}.
\newblock In \emph{ICLR}.

\bibitem[{Dai et~al.(2019)Dai, Chen, and Li}]{dai2019backdoor}
Jiazhu Dai, Chuanshuai Chen, and Yufeng Li. 2019.
\newblock A backdoor attack against lstm-based text classification systems.
\newblock \emph{IEEE Access}, 7:138872--138878.

\bibitem[{Devlin et~al.(2019)Devlin, Chang, Lee, and
  Toutanova}]{Devlin2019BERTPO}
J.~Devlin, Ming-Wei Chang, Kenton Lee, and Kristina Toutanova. 2019.
\newblock Bert: Pre-training of deep bidirectional transformers for language
  understanding.
\newblock In \emph{NAACL-HLT}.

\bibitem[{Gu et~al.(2017)Gu, Dolan-Gavitt, and Garg}]{Gu2017BadNetsIV}
Tianyu Gu, Brendan Dolan-Gavitt, and Siddharth Garg. 2017.
\newblock Badnets: Identifying vulnerabilities in the machine learning model
  supply chain.
\newblock \emph{ArXiv}, abs/1708.06733.

\bibitem[{Guan(2019)}]{Guan2019NeuralBI}
Andrew Guan. 2019.
\newblock Neural backdoors in nlp.

\bibitem[{Jean et~al.(2015)Jean, Cho, Memisevic, and Bengio}]{Jean2015OnUV}
S{\'e}bastien Jean, Kyunghyun Cho, Roland Memisevic, and Yoshua Bengio. 2015.
\newblock \href {https://doi.org/10.3115/v1/P15-1001} {On using very large
  target vocabulary for neural machine translation}.
\newblock In \emph{Proceedings of the 53rd Annual Meeting of the Association
  for Computational Linguistics and the 7th International Joint Conference on
  Natural Language Processing (Volume 1: Long Papers)}, pages 1--10, Beijing,
  China. Association for Computational Linguistics.

\bibitem[{Kalchbrenner and Blunsom(2013)}]{Kalchbrenner2013RecurrentCT}
Nal Kalchbrenner and P.~Blunsom. 2013.
\newblock Recurrent continuous translation models.
\newblock In \emph{EMNLP}.

\bibitem[{Kim(2014)}]{Kim2014ConvolutionalNN}
Yoon Kim. 2014.
\newblock Convolutional neural networks for sentence classification.
\newblock In \emph{EMNLP}.

\bibitem[{Kingma and Ba(2015)}]{kingma2014adam}
Diederik~P. Kingma and Jimmy Ba. 2015.
\newblock Adam: A method for stochastic optimization.
\newblock \emph{CoRR}, abs/1412.6980.

\bibitem[{Kurita et~al.(2020)Kurita, Michel, and Neubig}]{Kurita2020WeightPA}
Keita Kurita, Paul Michel, and Graham Neubig. 2020.
\newblock \href {https://doi.org/10.18653/v1/2020.acl-main.249} {Weight
  poisoning attacks on pretrained models}.
\newblock In \emph{Proceedings of the 58th Annual Meeting of the Association
  for Computational Linguistics}, pages 2793--2806, Online. Association for
  Computational Linguistics.

\bibitem[{Liu et~al.(2017{\natexlab{a}})Liu, Ma, Aafer, Lee, Zhai, Wang, and
  Zhang}]{liu2017trojaning}
Yingqi Liu, Shiqing Ma, Yousra Aafer, Wen-Chuan Lee, Juan Zhai, Weihang Wang,
  and Xiangyu Zhang. 2017{\natexlab{a}}.
\newblock Trojaning attack on neural networks.

\bibitem[{Liu et~al.(2017{\natexlab{b}})Liu, Xie, and
  Srivastava}]{liu2017neural}
Yuntao Liu, Yang Xie, and Ankur Srivastava. 2017{\natexlab{b}}.
\newblock Neural trojans.
\newblock In \emph{2017 IEEE International Conference on Computer Design
  (ICCD)}, pages 45--48. IEEE.

\bibitem[{Maas et~al.(2011)Maas, Daly, Pham, Huang, Ng, and
  Potts}]{Maas2011LearningWV}
Andrew~L. Maas, Raymond~E. Daly, P.~T. Pham, D.~Huang, A.~Ng, and Christopher
  Potts. 2011.
\newblock Learning word vectors for sentiment analysis.
\newblock In \emph{ACL}.

\bibitem[{Mu{\~n}oz-Gonz{\'a}lez et~al.(2017)Mu{\~n}oz-Gonz{\'a}lez, Biggio,
  Demontis, Paudice, Wongrassamee, Lupu, and Roli}]{munoz2017towards}
Luis Mu{\~n}oz-Gonz{\'a}lez, Battista Biggio, Ambra Demontis, Andrea Paudice,
  Vasin Wongrassamee, Emil~C Lupu, and Fabio Roli. 2017.
\newblock Towards poisoning of deep learning algorithms with back-gradient
  optimization.
\newblock In \emph{Proceedings of the 10th ACM Workshop on Artificial
  Intelligence and Security}, pages 27--38.

\bibitem[{Newell et~al.(2014)Newell, Potharaju, Xiang, and
  Nita-Rotaru}]{newell2014practicality}
Andrew Newell, Rahul Potharaju, Luojie Xiang, and Cristina Nita-Rotaru. 2014.
\newblock On the practicality of integrity attacks on document-level sentiment
  analysis.
\newblock In \emph{Proceedings of the 2014 Workshop on Artificial Intelligent
  and Security Workshop}, pages 83--93.

\bibitem[{Qi et~al.(2020)Qi, Chen, Li, Liu, and Sun}]{Qi2020ONIONAS}
Fanchao Qi, Yangyi Chen, Mukai Li, Zhiyuan Liu, and Maosong Sun. 2020.
\newblock Onion: A simple and effective defense against textual backdoor
  attacks.
\newblock \emph{ArXiv}, abs/2011.10369.

\bibitem[{Radford et~al.(2019)Radford, Wu, Child, Luan, Amodei, and
  Sutskever}]{Radford2019LanguageMA}
A.~Radford, Jeffrey Wu, R.~Child, David Luan, Dario Amodei, and Ilya Sutskever.
  2019.
\newblock Language models are unsupervised multitask learners.

\bibitem[{Shafahi et~al.(2018)Shafahi, Huang, Najibi, Suciu, Studer, Dumitras,
  and Goldstein}]{shafahi2018poison}
A.~Shafahi, W.~R. Huang, Mahyar Najibi, Octavian Suciu, Christoph Studer,
  T.~Dumitras, and T.~Goldstein. 2018.
\newblock Poison frogs! targeted clean-label poisoning attacks on neural
  networks.
\newblock In \emph{NeurIPS}.

\bibitem[{Socher et~al.(2013)Socher, Perelygin, Wu, Chuang, Manning, Ng, and
  Potts}]{Socher2013RecursiveDM}
R.~Socher, Alex Perelygin, J.~Wu, Jason Chuang, Christopher~D. Manning, A.~Ng,
  and Christopher Potts. 2013.
\newblock Recursive deep models for semantic compositionality over a sentiment
  treebank.
\newblock In \emph{EMNLP}.

\bibitem[{Turner et~al.(2018)Turner, Tsipras, and
  Madry}]{Turner2018CleanLabelBA}
Alexander Turner, D.~Tsipras, and A.~Madry. 2018.
\newblock Clean-label backdoor attacks.

\bibitem[{Yang et~al.(2019)Yang, Dai, Yang, Carbonell, Salakhutdinov, and
  Le}]{yang2019xlnet}
Zhilin Yang, Zihang Dai, Yiming Yang, J.~Carbonell, R.~Salakhutdinov, and
  Quoc~V. Le. 2019.
\newblock Xlnet: Generalized autoregressive pretraining for language
  understanding.
\newblock In \emph{NeurIPS}.

\bibitem[{Zhang et~al.(2015)Zhang, Zhao, and LeCun}]{Zhang2015CharacterlevelCN}
Xiang Zhang, Junbo Zhao, and Yann LeCun. 2015.
\newblock Character-level convolutional networks for text classification.
\newblock \emph{Advances in neural information processing systems},
  28:649--657.

\bibitem[{Zhang et~al.(2020)Zhang, Zhang, and Wang}]{Zhang2020TrojaningLM}
Xinyang Zhang, Zheng Zhang, and Tianying Wang. 2020.
\newblock Trojaning language models for fun and profit.
\newblock \emph{ArXiv}, abs/2008.00312.

\end{thebibliography}
\bibliographystyle{acl_natbib}

\newpage

\end{document}